\def\year{2022}\relax
\documentclass[letterpaper]{article} 
\usepackage{aaai22}  
\usepackage{times}  
\usepackage{helvet}  
\usepackage{courier}  
\usepackage[hyphens]{url}  
\usepackage{graphicx} 
\def\UrlFont{\rm}  
\usepackage{natbib}  
\usepackage{caption} 
\DeclareCaptionStyle{ruled}{labelfont=normalfont,labelsep=colon,strut=off} 
\usepackage{algorithm}
\usepackage{algorithmic}

\usepackage{amsmath}
\usepackage{amssymb}
\usepackage{bbm}
\usepackage{tabularx}
\usepackage{subfigure}

\DeclareMathOperator*{\argmin}{arg\,min}
\usepackage{newfloat}
\usepackage{listings}
\floatstyle{ruled}
\newfloat{listing}{tb}{lst}{}
\floatname{listing}{Listing}
\title{Towards Federated Clustering: A Federated Fuzzy $c$-Means Algorithm (FFCM)}
\author{
    Morris Stallmann\equalcontrib,
    Anna Wilbik\equalcontrib
}
\affiliations{
    Department of Data Science and Knowledge Engineering (DKE), Maastricht University\\
    Paul-Henri Spaaklaan 1, 6229 EN, Maastricht, The Netherlands\\
    m.stallmann@maastrichtuniversity.nl, a.wilbik@maastrichtuniversity.nl

}

\begin{document}

\maketitle

\begin{abstract}
Federated Learning (FL) is a setting where multiple parties with distributed data collaborate in training a joint Machine Learning (ML) model while keeping all data local at the parties. 
Federated clustering is an area of research within FL that is concerned with grouping together data that is globally similar while keeping all data local. We describe how this area of research can be of interest in itself, or how it helps addressing issues like non-independently-identically-distributed (i.i.d.) data in supervised FL frameworks. The focus of this work, however, is an extension of the federated fuzzy $c$-means algorithm to the FL setting (FFCM) as a contribution towards federated clustering. We propose two methods to calculate global cluster centers and evaluate their behaviour through challenging numerical experiments. We observe that one of the methods is able to identify good global clusters even in challenging scenarios, but also acknowledge that many challenges remain open.
\end{abstract}

\section{Introduction}
The success of Machine Learning (ML) can partly be attributed to the availability of good and sufficiently sized training datasets. Often, the data are stored on a central server, where ML models are trained. However, the data might initially be distributed among many clients (e.g., smartphones, companies, etc.) and gathering the data on a central server is not always feasible due to privacy regulations (like GDPR) \cite{gdpr}), the amount of data, or other reasons. Federated Learning (FL) is an approach that allows clients to jointly learn ML models while keeping all data local \cite{kairouz2021}.
Authors describe the generic FL training process by five steps:
\begin{enumerate}
    \item Client Selection: Select clients participating in the training.
    \item Broadcast: A central server initializes a global model and shares it with the clients.
    \item Client Computation: Each client updates the global model by applying a training protocol and shares the updates with the central server.
    \item Aggregation: The central server applies an aggregation function to update the global model.
    \item Model update: The updated global model is shared with the clients.
\end{enumerate}
This protocol can be repeated multiple times until a convergence criterion is met. Training a model following such a protocol has been successfully applied to a variety of use-cases, e.g., for next-word predictions on smartphones \cite{hard2018federated}, vehicle image classification \cite{dongdong_2020}, data collaboration in the healthcare industry \cite{deist2017infrastructure, BRISIMI_2018_health}, on IoT-data \cite{grefen2018complex, Moming_iot, wang_rl_2019}, and many more. For comprehensive surveys please refer to \cite{kairouz2021}, \cite{xuefei2021} or \cite{latif2021}. Many works focus on supervised learning while the area of federated unsupervised learning, like federated clustering, is less explored. 

Federated clustering is a FL setting, where the goal is to group together (local) datapoints that are globally similar to each other. That is, datapoints are distributed among multiple clients and are clustered based on a global similarity measure while all data remains local on client devices.
To the best of our knowledge, there are only few works addressing this problem (Section Related Work). We contribute to this field of research by introducing an extension of the fuzzy $c$-means algorithm to the FL setting. Fuzzy $c$-means is a clustering algorithm that allows data points to be assigned to more than one cluster (see Section Related Work). Thanks to its simplicity, it is widely used when such fuzzy assignments are required and we widen the area of application to include FL settings.

In particular, in this work, we describe a protocol that fits into steps described above, propose two different aggregation functions (straight-forward federated averaging and $k$-means averaging), and evaluate the federated fuzzy $c$-means algorithm on different datasets. We identify short-comings of the straight-forward aggregation and conclude that the $k$-means averaging is more suited to deal with non-i.i.d. data, but comes at a higher computational cost. Further, we acknowledge that many challenges (like proving the effectiveness outside a lab environment, determining the number of clusters in a federated setting, and others) remain open, but hope to inspire practitioners to experiment with unsupervised FL through the simplicity of our framework.

In the remainder of this section, we motivate research in federated clustering by describing two potential application cases.

\subsection{Motivational Examples}
We want to further motivate our work by describing two situations in which federated clustering could be applied. The first one illustrates how federated clustering can be of interest in itself while the second example shows how it could help to improve supervised FL. 

\begin{figure}[t!]
    \centering
    \subfigure[]{\includegraphics[width=0.22\textwidth]{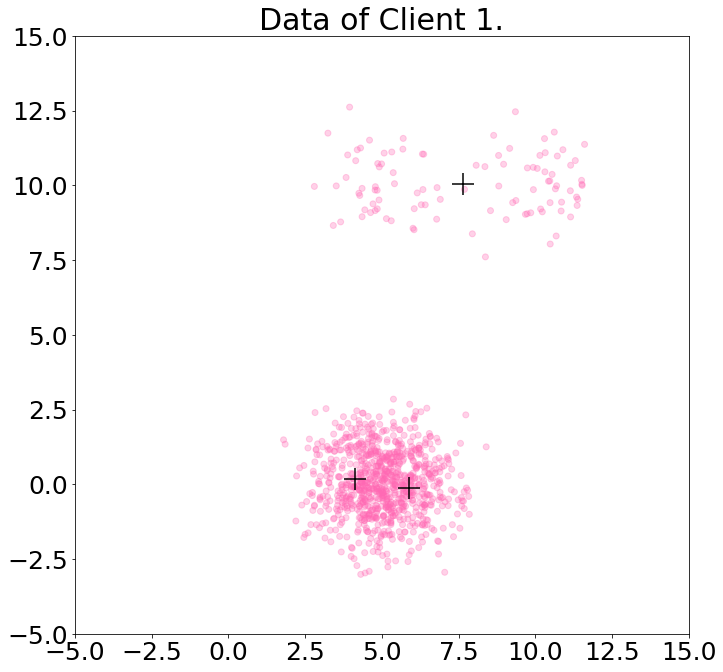}}
    \subfigure[]{\includegraphics[width=0.22\textwidth]{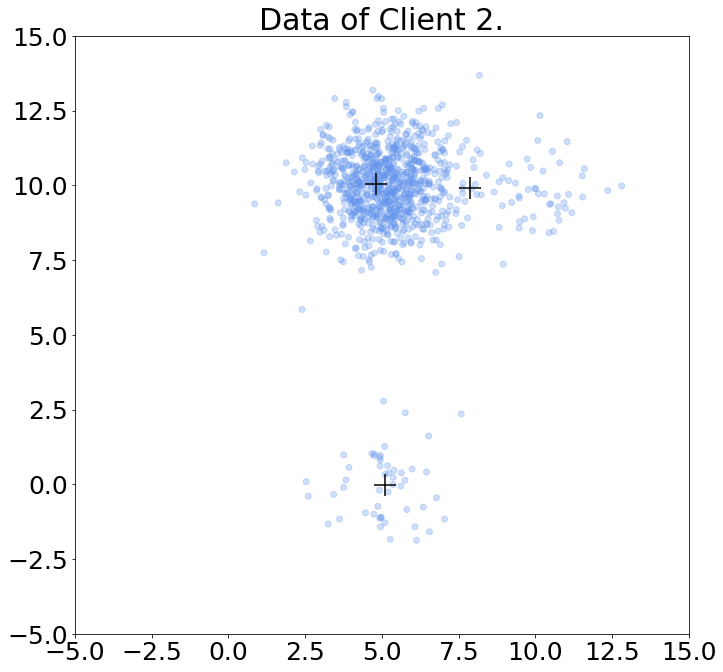}}
    \subfigure[]{\includegraphics[width=0.22\textwidth]{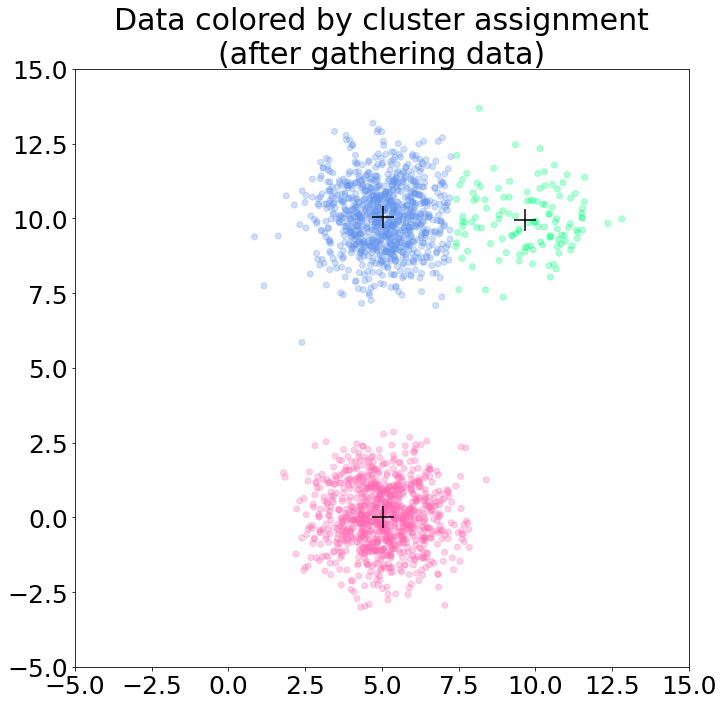}}
    \caption{Example for a situation where data collaboration is beneficial for all clients. (a) Client 1's local data. (b) Client 2's local data. (c) Data combined in a central storage. Only now all three cluster centers are identified.}
    \label{fig:motivational_example}
\end{figure}

Firstly, we start with an example to demonstrate how data-collaborators can benefit from federated clustering directly. Imagine a multi-national company with several local markets selling similar consumer goods in all markets. Each local market has data about their customers (e.g., age, place of residency, sold good, etc.) and applies clustering algorithms to generate customer segments in order to steer marketing activities. The company wishes to derive global clusters to understand their global customer base and identify unlocked potential in the local markets. Due to strict privacy regulations, the company is not allowed to gather all data in a central database (e.g., European customer data is not allowed to be transferred to most countries outside of Europe). The company could ask each local market to share their local cluster centers, but this approach disregards that clusters might only become apparent when the data is combined. For an example of such a situation, please refer to Figure \ref{fig:motivational_example}. In this artificial example, we know there exist three global clusters, because we generated them from three Gaussian distributions. The data are distributed among two clients. When each client is asked to identify three cluster centers from their local data, it results in two centers that are close to each other and one further away for both clients. They might reasonably conclude there exist only two clusters. However, when the data is combined, the fuzzy $c$-means algorithm is able to find all three cluster centers. With federated clustering, we aim to find the same global clusters without gathering all data in a central storage. In fact, we will see that this is possible by applying FFCM.

Secondly, federated clustering can be embedded into supervised FL to boost model performance. In many FL applications, the challenge of non-i.i.d. data makes learning a global model hard \cite{zhu2021federated, li2019survey}. Each client might have data from a different distribution. Many works have demonstrated the effectiveness of clustering clients together and utilize that clustering for more focused model updates (see Section Related Work) in such situations.
However, clients themselves might also have non-i.i.d. data (like in the example above). In such situations, assigning datapoints to clusters (rather than clients to clusters) could help to improve model performance similarly. In fact, authors of \cite{Caldarola_2021_CVPR} observe that the incorporation of domain (= cluster) information can improve a supervised system's performance. They incorporate (fuzzy) domain assignments into a convoluted and convolutional neural network architecture for image classification. In their experiments, they observe (slight) gain in accuracy as compared to not using the domain information.
For another example, assume the following situation: Some companies would like to collaboratively learn a model for product recommendation, but are not allowed to exchange data directly. Different clusters of clients might have very different needs (e.g., students and pensioners), and one recommender model for all might perform poorly. As described above (and in Figure \ref{fig:motivational_example}), some clusters might even only appear when the data is combined. One model per cluster has the potential to better address the different needs of customer segments. 

These two aspects let us conclude that it is worth investigating federated clustering as a research area of interest in itself. 
With this work, we introduce a federated fuzzy $c$-means algorithm and investigate how it behaves in different scenarios through numerical experiments. We hope to add new insights to the still new field of unsupervised FL and to open up new opportunities for improving existing supervised FL algorithms. 

\section{Background: Related Work}
\label{sec:relwork}
\subsection{Federated Learning}

In \cite{kairouz2021}, federated learning is defined as ``a machine learning setting where multiple entities (clients) collaborate in solving a machine learning problem, under the coordination of a central server or service provider. Each client’s raw data is stored locally and not exchanged or transferred; instead, focused updates intended for immediate aggregation are used to achieve the learning objective''.

In some situations, a single party might not have sufficient data to train a sufficiently good ML model, but multiple parties could train a good model if they combined their data. However, due to a multitude of reasons, this might not be possible or desirable. Federated learning offers an alternative by sharing model updates with a central server rather than the raw data. Hence, it opens up new potential for data collaborations and use-cases.
The field of research is still relatively new and many open challenges remain, e.g., efficient communication, data leakage through model updates, guaranteeing no party has an unfair advantage over others, and more. Please refer to \cite{kairouz2021} for a comprehensive study. 

There exist several works that adapt different ML algorithms to the FL setting. For example, regression models like in \cite{mcmahan2017communication,li2020practical,wang2021privacy}), classification algorithms like in \cite{BRISIMI_2018_health, dongdong_2020, bakopoulou2019federated, ying_multi-task_2019, Moming_iot}, or reinforcement learning \cite{wang_rl_2019}. The list is by no means exhaustive, but gives an impression of the variety of challenges addressed by FL.
Federated clustering is one more of these many challenges, and we discuss some works in that area in the next section.

\subsection{Federated Clustering}
Due to the similarity in terminology, we start by contrasting clustered federation with federated clustering. Clustered federation is concerned with identifying clusters of clients or model updates that are suitable to be grouped for a focused update of global supervised FL models. 
It has been proven to be effective when addressing issues caused by non-i.i.d. data among clients \cite{ghosh_clustered_federation, sattler_clustered_2020, kim_dynamic_clustering_2021, xie2021multi}.

On contrast, federated clustering is concerned with identifying global clusters in distributed data without sharing the data and, to the best of our knowledge, has not been explored as much. 
In \cite{kumar_fed_kmeans_2020}, the $k$-means algorithm was extended to the federated setting. They propose a global averaging function that calculates a weighted mean of local cluster centers in order to update global cluster centers, where the weights are given by the number of local datapoints assigned to the clusters. Further, there also exists a fuzzy version introduced in \cite{Pedrycz2021} that uses fuzzy assignments as weights instead of number of datapoints given by the hard assignments. Despite the similarity between \cite{Pedrycz2021} and part of our work, both were developed independently as we were not aware of \cite{Pedrycz2021} when we started. In fact, \cite{kumar_fed_kmeans_2020} and \cite{Pedrycz2021} are similar to our $avg_{1}$-function (Equation (\ref{eq:normal_avg})) that is explained in more detail in subsequent sections.
Both works show that this approach can identify reasonable clusters in some distributed datasets and we come to a similar conclusion in our experiments (see G2-sets experiments). However, we also identify scenarios in which a different aggregation function (Equation (\ref{eq:k-means_avg})) results in arguably better clusters (see hidden clusters and locally absent cluster experiments). This aggregation function applies a $k$-means clustering algorithm to the clients' local centers. At the time of writing the initial version of this paper, we were not aware of any work in federated clustering applying a similar averaging function. However, in the meanwhile, \cite{pmlr-v139-dennis21a} independently introduced the application of a $k$-means averaging in the context of federated one-shot clustering with $k$-means. The one-shot clustering, however, follows a different protocol. Our version applies multiple rounds of communication to detect even ``hidden" clusters (see Figure \ref{fig:motivational_example} and hidden cluster experiments). The one-shot version minimizes the communication overhead and is not designed to detect such hidden clusters centers.
Another work we would like to highlight is \cite{Caldarola_2021_CVPR}, because it shows how a supervised FL framework can benefit from federated clustering. The authors identify different clusters in clients' local data by applying a student-teacher paradigm. These clusters are then embedded into a supervised FL framework. Authors conclude that incorporating the cluster information can lead to better performance of the supervised system.  

\subsection{Non-Federated Clustering via $k$-Means and Fuzzy $c$-Means}
\begin{table}[t!]
\centering
\begin{tabularx}{0.4\textwidth}{X|X}
\textbf{Symbol} & \textbf{Meaning} \\ \hline
$X=[x_{i}\in\mathbb{R}^{d}]_{i=1}^{N}$ & $\mathbb{N}\ni d$-dimensional data points to be clustered\\
 $c=[c_{j}]_{j=1}^{K}$ & Center of cluster $j=1,\dots, K$ \\
$U=[u_{i,j}]_{i=1}^{N}, j=1, \dots K$ & Membership of data point $i$ to cluster $j$; partition of data $X$ \\ 
 $1\le m<\infty$ & weighting exponent to control fuzziness \\ 
\end{tabularx}
\caption{Symbols used in the (federated) fuzzy $c$-means and $k$-means formulations.}
\label{tab:fuzzy_cmeans_symbols}
\end{table}

In our proposed method, the $k$-means and fuzzy $c$-means algorithms play an important role. Therefore, we remind the reader of these two algorithms. In subsequent sections, we are going to use notation as listed in Table \ref{tab:fuzzy_cmeans_symbols}.

\subsubsection{$k$-Means}
Let $X$ be a given data-set. The objective of the $k$-means clustering is to find cluster centers $c=(c_{1}, c_{2}, \dots, c_{K})$ (and corresponding assignments) such that the following expression is minimized (see e.g. \cite{Hastie17theelements_kmeans}):
\begin{align}
    \hat{J}(c) &= \sum_{k=1}^{K} \sum_{x\in X} I(x, k) ||x-c_{k}||^{2} \\
    I(x, k) &= \begin{cases} 1, x \text{ is assigned to cluster }k, \nonumber \\
    0,\text{ otherwise}.
    \end{cases}
\end{align}
Each datapoint $x$ is assigned to its closest cluster center, where closeness is defined by euclidean distance. The assignment is given by function $a:X\times \{c_{1}, \dots, c_{K}\} \to \{1, \dots, K\}$:
\begin{align}\label{def:k_means_assignment}
        a(x, \{ c_{k} \}_{k=1}^{K}) := \argmin_{1\le k \le K} ||c_{k}-x||^{2}
\end{align}

A widely-used iterative algorithm for finding such a clustering works as follows \cite{Hastie17theelements_kmeans}:
\begin{enumerate}
    \item Initialize cluster centers $c_{1}, \dots, c_{K}$.
    \item Assign clusters for all $x\in X$ according to $a(x, \{ c_{k} \}_{k=1}^{K})$ (Definition (\ref{def:k_means_assignment})).
    \item Recalculate cluster centers $c_{k}$ by solving the following problem, i.e., calculating the mean of points assigned to the same cluster:
    \begin{align}
        c_{k} = \min_{m\in\mathbb{R}^{d}}\sum_{x\in X:a(x)=k} || x- m ||^{2}, k=1,\dots, K. \nonumber
    \end{align}
    \item Check if the convergence criterion is met, i.e., whether the assignment did not change (much) compared to the previous iteration. Let $a_{i}^{(t-1)}$ be the assignment vector of the previous iteration for datapoint $x_{i}\in X$, i.e., the $k$-th entry is $1$ if $a(x_{i}, \{ c_{k} \})=k$ and zero otherwise. Let $a_{i}^{(t)}$ be the assignment of the current iteration. Further, let $A^{(t-1)}$ and $A^{(t)}$ be the assignment matrices, where the $i$-th row equals $a_{i}^{(t-1)}$ and $a_{i}^{(t)}$, respectively. Then, the algorithm converged if the difference between the two assignment matrices is smaller than some predefined $\epsilon$:
    \begin{align}
        ||A^{(t-1)}-A^{(t)}||^{2}<\epsilon.
    \end{align}
    If the algorithm did not converge yet, go back to step 2. If it did converge, terminate.
\end{enumerate}
$\hat{J}(c)$ is monotonously decreasing with each iteration, but it is known that the algorithm might get stuck in a local minimum. In fact, it does not offer any performance guarantee and \cite{arthur2007_k++} argues it often fails due to its sensitivity to the initialization method. In \cite{arthur2007_k++}, the still popular initialization method $k$-means++ is introduced. It subsequently chooses random cluster centers that are likely to be far from already chosen centers.
In our experiments, we use the scikit-learn implementation that applies the $k$-means++ initialization method, too \cite{scikit-learn}.

\subsubsection{Non-Federated Fuzzy $c$- Means}\label{non-fed-means}
Fuzzy $c$-means is a well-known soft clustering method that assigns a membership index $u_{ij}$ for clusters $j=1,\dots, K$ to data points $x_{i}\in X$ such that
    $\sum_{j}^{K} u_{ij}=1\;\forall i=1, \dots, N$.
The term soft clustering refers to the fact that points are allowed to belong to more than one cluster. On contrast, a hard clustering method like $k$-means assigns each point to exactly one cluster.

For given data $X$, the objective is to find cluster centers $c_{j}$ and membership matrix $U=[u_{ij}]$ such that the following expression is minimized \cite{BEZDEK1984}:
\begin{align}
    J_{m}(U,c) &= \sum_{i=1}^{N}\sum_{j=1}^{K} (u_{ij})^{m} ||x_{i}-c_{j}||^{2}, \label{eq:objective_function} \\
    u_{ij} &:= \frac{1}{\sum_{k=1}^{K}(\frac{||x_{i}-c_{j}||^{2}}{||x_{i}-c_{k}||^{2}})^{\frac{2}{m-1}}}, \label{eq:fuzzy_member_update}
\end{align}
where $||y|| := \sqrt{\sum_{l=1}^{n} y_{l}^2}$.
It is closely related to the $k$-means clustering, and the main difference lies in the assignment matrices $A$ in $k$-means versus $U$ in fuzzy $c$-means.

Parameter $m>1$ controls how fuzzy the cluster assignments should be. The greater $m$, the more fuzziness in the assignment, i.e., points are assigned to more clusters with smaller values. A common choice that we also employ is $m=2$.

A widely used algorithm to find a solution to the optimization problem was introduced by \cite{BEZDEK1984} and follows four basic steps:

\begin{enumerate}
    \item Initialize matrix $U$ := $U^{0}$.
    \item In iteration $t$, (re)calculate cluster centers $c_{j}$ according to:
    \begin{align}\label{eq:center_update}
        c_{j} = \frac{\sum_{i} u_{ij}^{m} x_{i}}{\sum_{i}u_{ij}^{m}}
    \end{align}
    \item Update Membership Matrix $U^{t+1}$ according to Equation (\ref{eq:fuzzy_member_update}).
    \item Check if the convergence criterion is met: $||U^{t+1}-U^{t}||\le\epsilon$ for some predefined $\epsilon$, i.e., did the memberships change by at most $\epsilon$. If it was not met, return to step 2 after setting $U^{t}=U^{t+1}$. Terminate if it was met.
\end{enumerate}
The time-complexity of the algorithm is quadratic in the number of clusters $K$, and methods to reduce the complexity have been proposed \cite{kolen_reducing_2002}. With today's computational capabilities, even the original version can be applied to fairly big datasets, however. Similar to $k$-means, other short-comings of the algorithm are sensitivity to the cluster initialization and sensitivity to noise as noted in \cite{Suganya2012FuzzyCM}.
Those challenges have been addressed by subsequent works, but each auxiliary method comes with its own short-comings \cite{Suganya2012FuzzyCM}. It remains up to the practitioners to decide on a suitable method for their specific problems. 

For the introduction of federated fuzzy $c$-means, we focus on the original formulation and generalize it to the federated setting.

\section{Federated Fuzzy $c$-Means (FFCM)}
Our proposed method is an extension of the iterative fuzzy $c$-means algorithm to the federated learning setting similar to \cite{Pedrycz2021} and \cite{kumar_fed_kmeans_2020}, but with a different take on the global cluster center calculation.

In our scenario, the data is not stored in a centralized database, but distributed among multiple clients. The goal is to learn a global clustering that is similar to the clustering of the centralized data while the data stays private. The general procedure is as follows: Each client runs a number of fuzzy $c$-means iterations locally, and sends the resulting cluster centers to a central server. The central server is responsible for calculating meaningful global clusters from the local learners' results. After calculating the global centers, they are shared with the clients that use them to recalculate their local centers, which in turn are shared with the central server, and so forth. That procedure is repeated until the global centers remain stable.

The creation of a global model from clients' local model updates was first introduced by \cite{mcmahan2017communication} and is known as \emph{Federated Averaging} (FedAvg). We apply two averaging methods, where one of them is closely related to the ones in \cite{kumar_fed_kmeans_2020, Pedrycz2021} and the other one is a $k$-means averaging that was independently developed and applied in the context of federated one-shot clustering \cite{pmlr-v139-dennis21a}.

In the previous section, it is assumed that all data $X$ is stored centrally. Now, assume, data $X$ is distributed among $P$ parties (clients), i.e., $X = \bigcup_{l=1}^{P} X^{(l)}$.
Further, assume, each party's data $X^{(l)}$ is constrained to remain private. 

Therefore, we introduce a reformulation of non-federated fuzzy $c$-means that removes the necessity of storing all data centrally and, instead, fits into the Federated Learning framework:
\begin{enumerate}
    \item The central server initializes $K$ global cluster centers $c_{1}, \dots, c_{K}$.
    \item The central server shares $c_{1}, \dots, c_{K}$ with the clients.
    \item Client $l$ calculates membership matrix $U_{l}^{t+1}$ according to Equation (\ref{eq:fuzzy_member_update}) and generates local cluster centers $c_{1}^{(l)}, \dots, c_{K}^{(l)}$ according to Equation (\ref{eq:center_update}) ($l=1, \dots, P$).
    \item Client $l$ shares $c_{1}^{(l)}, \dots, c_{K}^{(l)}$ and each cluster's support or weight
    \begin{align}\label{eq:client_cluster_weight}
        \sum_{i}(u_{ij}^{(l)})^{m}=:W_{j}^{(l)},
    \end{align}
    for $j=1, \dots, K$ and $l=1,\dots, P$.
    \item The central server updates $c_{1}, \dots, c_{K}$ by applying an averaging function $avg([c_{1}^{(l)}, \dots, c_{K}^{(l)}]_{l=1}^{P}, [W_{j}^{(l)}]_{l=1}^{P})$.
    \item The central server checks a convergence criterion. If not converged, go back to step 2.
\end{enumerate}
Since the central server has only access to the local cluster centers $c_{k}^{(l)}$ and client cluster weights $W_{k}^{(l)}$, the previous convergence criterion can not be applied. As an alternative, we check whether the cluster centers changed by less than $\epsilon$ between two iterations. Let $c_{k_t}$ be the global cluster center $k$ after time step $t$. Then, the convergence criterion can be formulated as follows:
    $\sum_{k=1}^{K}||c_{k_t} - c_{k_{t+1}}|| \le \epsilon$.
Note that this new criterion might lead to different cluster centers than in the previous formulation. It lets the center move closer to the center of mass even though the assignments might have stabilized already.

In order to find meaningful global clusters, it is essential to find a good averaging function $avg(\cdot)$ used in step 5 of the framework outlined above. In this work, we propose two alternatives and evaluate them in our experiments.

The first global averaging function takes all locally calculated cluster centers and their respective weights $W_{j}^{(l)}$ as input. It calculates the global cluster centers by a weighted mean of the local cluster centers:
\begin{align}\label{eq:normal_avg}
    avg_{1}:\mathbb{R}^{P\times d}\times \mathbb{R}^{P}  &\to \mathbb{R}^{d} \nonumber \\
    avg_{1}([c_{k}^{(l)}], [W_{k}^{(l)}]) & :=  \frac{\sum_{l=1}^{P}W_{k}^{(l)}c_{k}^{(l)}}{\sum_{l=1}^{P}W_{k}^{(l)}} \\
    & = c_{k}. \nonumber
\end{align}
As we will see in subsequent sections, $avg_{1}$ can result in undesired results if the data is unequally distributed among clients. 
To address this problem, we introduce a second averaging function:
\begin{align}\label{eq:k-means_avg}
    avg_{2}:\mathbb{R}^{P\times d\times K} & \to \mathbb{R}^{d\times K} \nonumber \\
    avg_{2}([c_{k}^{(l)}]_{k,l}) & := kmeans([c_{k}^{(l)}]_{k,l}), \\
    & = [c_{k}]_{k}, \nonumber
\end{align}
where $kmeans(\cdot)$ denotes a function that applies the $k$-means clustering algorithm and outputs the $k$ cluster centers it converged to. Note, that we shorten notation in above Equation to improve readability.
This second averaging function applies the $k$-means algorithm to all reported local cluster centers to find new global cluster centers. 
It does introduce increased complexity, but the resulting cluster matching (that now is independent from weights assigned by clients) shows to be more robust against unequally distributed data in our experiments.

\section{Numerical Experiments}\label{sec:experiments}
In order to test the effectiveness of the federated fuzzy $c$-means algorithm, we perform a series of numerical experiments on different datasets. Firstly, we construct datasets by hand that model challenging situations, e.g., overlapping, unequally distributed or locally absent clusters. Secondly, we evaluate the algorithm on benchmark datasets from the literature \cite{clust_basic_bench:2018}. In this second series of evaluation, we assume the data to be uniformly distributed among clients and study how the federated algorithm is affected by increasing dimensionality and overlap. 

Our goal with the federated fuzzy $c$-means algorithm is to find clusters at least as good as the ones from fuzzy $c$-means on the combined dataset (subsequently also referred to as ``gathered" or ``central clustering"). To compare the results, we introduce and report a knowledge-gap metric alongside the within and outside cluster sum of squared error (defined below).

In all of our experiments we choose $m=2$ and report results for the two averaging functions (Equations (\ref{eq:normal_avg}) and (\ref{eq:k-means_avg})). To account for randomness in drawing from the distributions and in the initialization, we repeat each experiment ten times and report the averages.
Note that we assume the number of clusters to be known and leave the challenge of determining that number in a federated setting for future works.

More thorough descriptions of the datasets and metrics can be found in subsequent sections.
\subsection{Evaluation Metrics}
A commonly used approach to compare clustering results is to compare the sum of squared error (SSE). It is based on the intuition that points belonging to the same cluster should be similar. The distances between each data point and its assigned cluster's center (we translate the fuzzy assignments by taking the $\max$ of the weights) are summed to form an index for cluster cohesion, sometimes normalized by the number of datapoints $N$ and dimension $d$:
    $WSSE := \frac{\sum_{k=1}^{K}\sum_{x_{j}\in C_{k}}||x_{j}-c_{k}||^{2}}{N*d}$.
We refer to $WSSE$ as ``Within SSE" in our experiments and say, the smaller the $WSSE$, the better the cluster cohesion.

Similarly, we can define the ``Outside SSE" as the sum over the difference between data points and the cluster centers it is not assigned to as a measure for cluster separation:

    $OSSE := \frac{\sum_{k=1}^{K}\sum_{x_{j}\notin C_{k}}||x_{j}-c_{k}||^{2}}{N*d}.$
We say, the bigger, $OSSE$, the better the cluster separation.

Additionally, we have the luxury of knowing ground truth centers in our experiments. We use that information to compare how well the non-federated and federated algorithms match those. We define the knowledge gap between two sets of cluster centers $\tilde{c}=(\tilde{c}_{1}, \dots, \tilde{c}_{K})$ and $c=(c_{1}, \dots, c_{K})$ as:
    $gap := \sum_{k=1}^{K}||\tilde{c}_{k}-c_{k}||$.
It might not always be necessary to exactly match the ground truth for good cluster assignments, but the knowledge gap gives a good indication on whether the algorithm converged to meaningful results.
Note that $gap$ naturally increases with the number of clusters and dimensionality of the data. That is why we report a normalized version for the G2-sets experiments that have different dimensionality for better comparability: $ngap:=\frac{gap}{\sqrt{d}}$.

\subsection{Test Data and Results}
In each of the following subsection, we introduce the datasets before reporting the results of our experiments on those datasets.

\subsubsection{Case 1: Verification of the Implementation and Sensitivity to Unequally Distributed Data}
\begin{figure}[t!]
    \centering
    \subfigure[]{\includegraphics[width=0.16\textwidth]{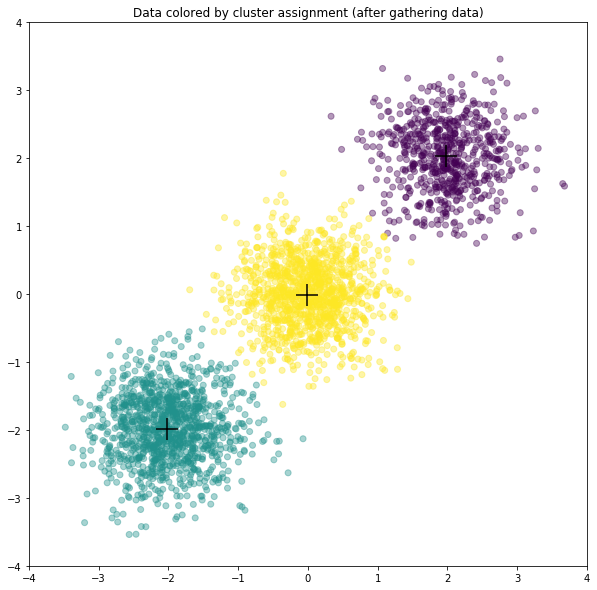}}
    \subfigure[]{\includegraphics[width=0.16\textwidth]{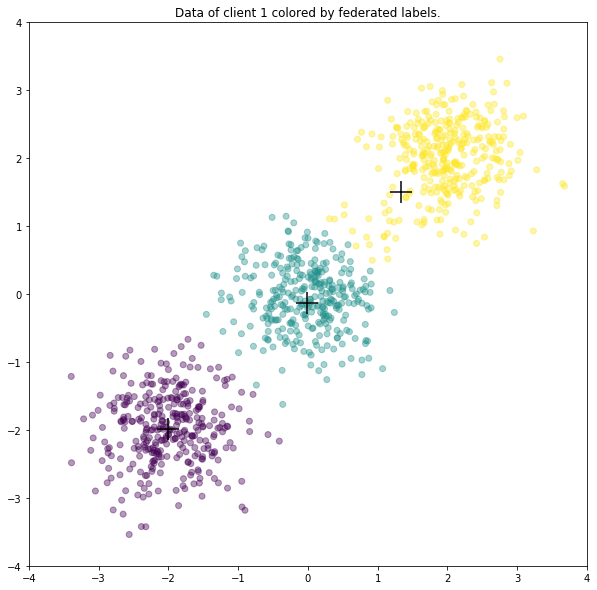}}
    \subfigure[]{\includegraphics[width=0.16\textwidth]{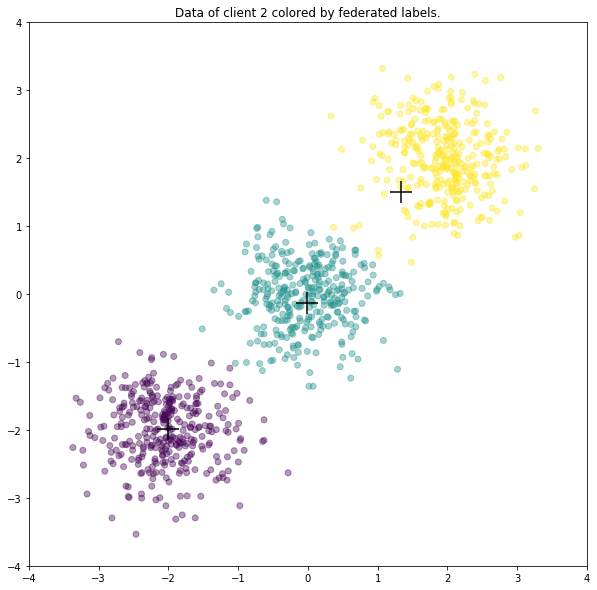}}
    \subfigure[]{\includegraphics[width=0.16\textwidth]{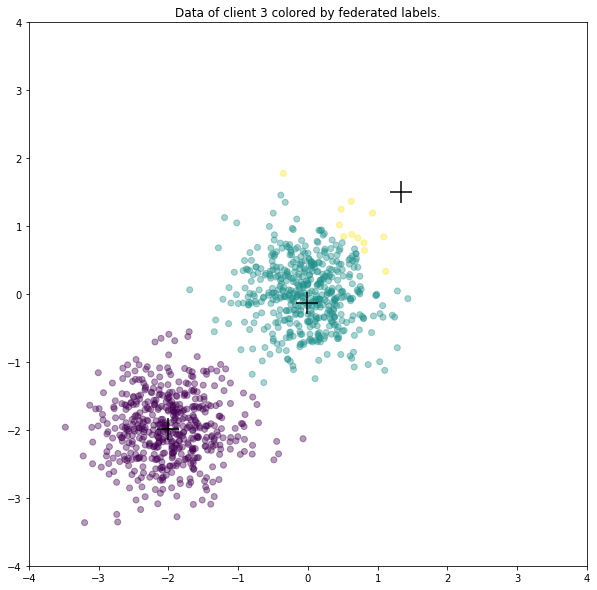}}
    \caption{Visualising test case 1 shows the sensitivity of $avg_{1}$ to unequally distributed points and motivates the introduction of $avg_{2}$. 
    (a) The ground truth as identified by clustering after combining the data (centers denoted by crosses). (b) Client 1's data with the centers as found by the federated clustering algorithm. (c) Client 2's data with the centers as found by the federated clustering algorithm. (d) Client 3's data with the centers as found by the federated clustering algorithm. }
    \label{fig:exp_verify}
\end{figure}

To verify the approach and correctness of our implementation, we start with a toy example. The dataset is composed of three 2-dimensional clusters, where the clusters are created by drawing from Gaussian distributions with means $\mu_{1}=(-2,-2)$, $\mu_{2}=(0,0)$, $\mu_{3}=(2,2)$. There are three clients with $999$ points each. Each local cluster has $333$ points drawn from one of the three distributions. The federated fuzzy $c$-means algorithm is able to detect the clusters like we expected.
In a second experiment, client 1 and client 2 have the same data while client 3 does not have any data drawn from the distribution with $\mu_{3}$, but $500$ points each from $\mu_{1}$ and $\mu_{2}$ distributions. In Figure \ref{fig:exp_verify} and Table \ref{tab:sensitivity_unequal_data}, the experiment and results are depicted. The centers are still close to the ground truth centers, but one of the center visibly deviates from the ground truth, that is found when combining the data. This hints to sensitivity of $avg_{1}$ to unequally distributed data, that we investigate in subsequent experiments and address by introducing $avg_{2}$. In fact, in this example, using $avg_{2}$ instead of $avg_{1}$ can improve the clustering (see Table \ref{tab:sensitivity_unequal_data}).

\begin{table}[t!]
\centering
\resizebox{0.48\textwidth}{!}{%
\begin{tabular}{l|ll|lll|lll}
 & \multicolumn{2}{l|}{\begin{tabular}[c]{@{}l@{}}Centrally\\ Clustered\end{tabular}} & \multicolumn{3}{l|}{$avg_{1}$} & \multicolumn{3}{l}{$avg_{2}$} \\ \hline
\begin{tabular}[c]{@{}l@{}}True \\ Centers\end{tabular} & \begin{tabular}[c]{@{}l@{}}Within \\ SSE\end{tabular} & \begin{tabular}[c]{@{}l@{}}Outside \\ SSE\end{tabular} & \begin{tabular}[c]{@{}l@{}}Fed.\\ Centers\end{tabular} & \begin{tabular}[c]{@{}l@{}}Within \\ SSE\\ (mean)\end{tabular} & \begin{tabular}[c]{@{}l@{}}Outside \\ SSE\\ (mean)\end{tabular} & \begin{tabular}[c]{@{}l@{}}Fed.\\ Centers\end{tabular} & \begin{tabular}[c]{@{}l@{}}Within \\ SSE\\ (mean)\end{tabular} & \begin{tabular}[c]{@{}l@{}}Outside \\ SSE\\ (mean)\end{tabular} \\ \hline
\begin{tabular}[c]{@{}l@{}}(-2,2), \\ (0,0),\\ (2,2)\end{tabular} & 0.3120 & 3.7070 & \begin{tabular}[c]{@{}l@{}}(-2.01, -1.98),\\ (-0.01, -0.13),\\ (1.34, 1.49)\end{tabular} & 0,3588 & 3,4078 & \begin{tabular}[c]{@{}l@{}}(-1.98, -1.97)\\ (0.00, -0.01),\\ (2.00, 2.00)\end{tabular} & 0.3120 & 3,7041
\end{tabular}%
}
\caption{In test case 1, applying $avg_{2}$ rather than $avg_{1}$ leads to centers that are closer to the ground truth and also improve within and outside cluster SSE metrics.}
\label{tab:sensitivity_unequal_data}
\end{table}

\subsubsection{Case 2: Hidden Clusters}
This test set addresses the motivational example in the introduction and is illustrated in Figure \ref{fig:motivational_example}. It is designed to illustrate the capability of our approach, but also show its limitations.
There are two clients with three clusters each. We know there are three clusters, because we generated them by drawing from three Gaussian distributions with means $\mu_{1}=(5,0)$, $\mu_{2}=(5,10)$, $\mu_{3}=(10,10)$, and standard deviation $\sigma =1.1$. We define $\mu_{i}\;(i=1,2,3)$ as our ground truth centers. Client 1 and and client 2 have $1000$ data points each. Client 1 has $900$ points drawn from the Gaussian with $\mu_{1}$ and $50$ points each from $\mu_{2},\mu_{3}$ while client 2 has $900$ points drawn from the distribution with $\mu_{2}$ and 50 each from the other distributions.
During our experiments, we noticed that the non-federated and federated versions struggle with identifying the three clusters. We tried different parameters $m$ and found that both versions work better with smaller values for $m$. Other than in the other sections, the hereby reported results are generated with $m=1.1$. 

Note that cluster center $\mu_{3}$ is only detected by (non-federated) fuzzy $c$-means if the data was combined first. Most of the time it was not detected by running the non-federated fuzzy $c$-means on either client's local data only in our experiments (see Figure \ref{fig:motivational_example} for illustration).

The results can be found in Table \ref{tab:hidden_clusters}. On the one hand side, we observe that by applying $avg_{2}$, the ground truth centers are found more reliably than by applying $avg_{1}$ (smaller ground truth gap). The ground truth gap is even smaller than in the central case. This can be explained by different convergence criteria that are checked in the formulations. In the central algorithm, it is checked whether the assignments stay stable. In the federated version it is checked whether the centers itself stay stable. This causes the centers to move closer to the center of mass.
On the other hand, we observe only slightly worse results for $avg_{1}$ in the SSE metrics.
All in all, we can conclude that $avg_{2}$ is more suited to identify hidden cluster centers than $avg_{1}$ (smaller ground truth gap), but also see that the cluster quality is only slightly improved. Moreover, we observe that choosing $m$ wisely is necessary to derive meaningful centers and acknowledge that more rigor work towards that direction is required.

\begin{table}[t!]
\centering
\resizebox{0.48\textwidth}{!}{%
\begin{tabular}{l|lll|lll|lll}
 & \multicolumn{3}{l|}{\begin{tabular}[c]{@{}l@{}}Centrally\\ Clustered\end{tabular}} & \multicolumn{3}{l|}{$avg_{1}$} & \multicolumn{3}{l}{$avg_{2}$} \\ \hline
\begin{tabular}[c]{@{}l@{}}Points \\ per\\ Client\end{tabular} & \begin{tabular}[c]{@{}l@{}}Ground\\ Truth \\ Gap\end{tabular} & \begin{tabular}[c]{@{}l@{}}Within\\ SSE\end{tabular} & \begin{tabular}[c]{@{}l@{}}Outside\\ SSE\end{tabular} & \begin{tabular}[c]{@{}l@{}}Ground\\ Truth\\ Gap\end{tabular} & \begin{tabular}[c]{@{}l@{}}Within\\ SSE\\ (mean)\end{tabular} & \begin{tabular}[c]{@{}l@{}}Outside\\ SSE\\ (mean)\end{tabular} & \begin{tabular}[c]{@{}l@{}}Ground \\ Truth \\ Gap\end{tabular} & \begin{tabular}[c]{@{}l@{}}Within\\ SSE\\ (mean)\end{tabular} & \begin{tabular}[c]{@{}l@{}}Outside\\ SSE\\ (mean)\end{tabular} \\ \hline
\begin{tabular}[c]{@{}l@{}}(1000,\\ 1000)\end{tabular} & 1.47 & 0.71 & 8.57 & 2.77 & 0.73 & 8.29 & 1.25 & 0.72 & 8.55 \\ \hline
\end{tabular}%
}
\caption{Experiments with the ``hidden cluster" (case 2) dataset: One of the clusters only becomes apparent when the data is combined and is (most of the time) not recognized when the non-federated fuzzy $c$-means is run on the local data only. In our experiments, we observe that the cluster centers can be found without gathering the data more reliably by applying $avg_{2}$ rather than $avg_{1}$ (smaller ground truth gap). SSE metrics for $avg_{1}$ are only slightly worse.}
\label{tab:hidden_clusters}
\end{table}

\subsubsection{Case 3: Overlapping and Locally Absent Clusters}
\begin{figure}[t!]
    \centering
    \includegraphics[width=0.3\textwidth]{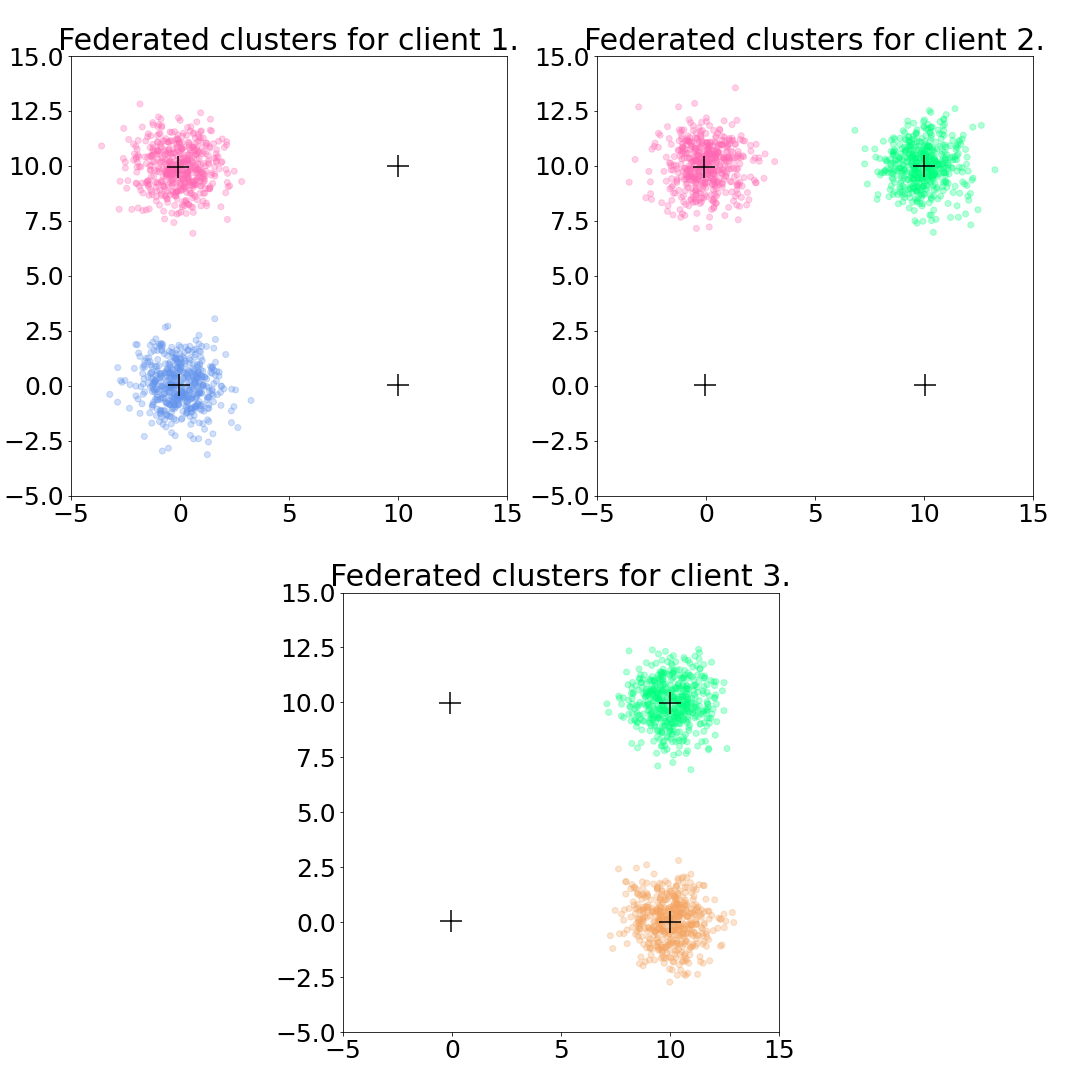}
    \caption{Data in the case 3 (``locally absent dataset''). The upper left figure shows client 1's local data, the upper right one client 2's local data and the lower one client 3's local data. Crosses denote global cluster centers as identified by federated clustering with $avg_{2}$.}
    \label{fig:locally_absent_clusters}
\end{figure}

The next set of training sets is designed to study how the algorithm performs in case of unequally distributed, locally absent, and overlapping clusters (see Figure \ref{fig:locally_absent_clusters} for illustration). There are three clients in total. Each client has two 2-dimensional clusters locally, and there are four global clusters in total. The clusters are generated by sampling from Gaussian distributions centered at $\mu_{1}=(0,0)$, $\mu_{2}=(0,10)$, $\mu_{3}=(10,10)$, and $\mu_{4}=(10,0)$ with standard deviation of $1.0$. Client 1 has data drawn from Gaussians with $\mu_{1}, \mu_{2}$, client 2 $\mu_{2}, \mu_{3}$ and client 3 $\mu_{3}, \mu_{4}$. In Figure \ref{fig:locally_absent_clusters}, the experimental setup is depicted.
We assume, each client's local clusters have equal size, but vary the relative number of local datapoints (i.e., in one experiment client 1 has most points in total, in another experiment client 2 has most points, etc.). 
Note that $avg_{1}$ (Equation (\ref{eq:normal_avg})) uses the cluster assignment weights reported by the local clients to update the global centers, which might be problematic in case the clusters are absent in some clients. No client alone can find four meaningful clusters, and, hence, the reported centers and according weights might not be meaningful such that the global learner might have difficulties in finding good global centers.

The results can be found in Table \ref{tab:locally_absent_clusters}.
Indeed, we observe again that $avg_{1}$ fails in situations where the clusters are unequally distributed among and absent in clients. However, we also observe that the $k$-means averaging $avg_{2}$ (Equation (\ref{eq:k-means_avg})) can better cope with such situations and leads to finding clusters that are close to the ground truth. This is also reflected in the SSE metrics that show better cluster cohesion and separation.

\begin{table}[t!]
\centering
\resizebox{0.48\textwidth}{!}{%
\begin{tabular}{l|lll|lll|lll}
 & \multicolumn{3}{l|}{\begin{tabular}[c]{@{}l@{}}Centrally\\ Clustered\end{tabular}} & \multicolumn{3}{l|}{$avg_{1}$} & \multicolumn{3}{l}{$avg_{2}$} \\ \hline
\begin{tabular}[c]{@{}l@{}}Points\\ per\\ Client\end{tabular} & \begin{tabular}[c]{@{}l@{}}Ground\\ Truth\\ Gap\end{tabular} & \begin{tabular}[c]{@{}l@{}}Within \\ SSE\end{tabular} & \begin{tabular}[c]{@{}l@{}}Outside\\ SSE\end{tabular} & \begin{tabular}[c]{@{}l@{}}Ground\\ Truth\\ Gap\end{tabular} & \begin{tabular}[c]{@{}l@{}}Within \\ SSE\\ (mean)\end{tabular} & \begin{tabular}[c]{@{}l@{}}Outside\\ SSE\\ (mean)\end{tabular} & \begin{tabular}[c]{@{}l@{}}Ground\\ Truth\\ Gap\end{tabular} & \begin{tabular}[c]{@{}l@{}}Within \\ SSE\\ (mean)\end{tabular} & \begin{tabular}[c]{@{}l@{}}Outside\\ SSE\\ (mean)\end{tabular} \\ \hline
\begin{tabular}[c]{@{}l@{}}(100,\\ 1000,\\ 100)\end{tabular} & 0.41 & 0.70 & 15.24 & 1.17 & 1.88 & 12.92 & 0.12 & 0.63 & 17.18 \\ \hline
\begin{tabular}[c]{@{}l@{}}(100, \\ 1000,\\ 1000)\end{tabular} & 0.21 & 0.65 & 16.00 & 3.88 & 1.28 & 11.95 & 0.08 & 0.63 & 17.21 \\ \hline
\begin{tabular}[c]{@{}l@{}}(1000,\\ 100,\\ 100)\end{tabular} & 0.17 & 0.65 & 16.55 & 1.34 & 2.28 & 13.83 & 0.10 & 0.63 & 17.18 \\ \hline
\begin{tabular}[c]{@{}l@{}}(1000, \\ 1000,\\ 1000)\end{tabular} & 0.04 & 0.63 & 17.14 & 3.62 & 1.16 & 11.69 & 0.03 & 0.62 & 17.14
\end{tabular}%
}
\caption{Experiments with the case 3 (``locally absent clusters") dataset: Through the knowledge gap metric, we observe that $avg_{2}$ is better suited to detect the ground truth centers than $avg_{1}$. This also leads to favorable results in the SSE metric.}
\label{tab:locally_absent_clusters}
\end{table}

\subsubsection{Case 4: G2 Sets}
Additionally, we test our method on more cluster benchmark sets from an online repository \cite{clust_basic_bench:2018}. In particular, the G2 sets were introduced in \cite{G2sets} and each set was generated by drawing $2048$ samples from two Gaussian distributions with different means, i.e. each set contains two ground truth centers. The Gaussians are centered at $\mu_{1}=(500, 500, \dots)$ and $\mu_{2}=(600, 600, \dots)$ with standard deviations $\sigma\in \{10, 20, \dots, 100 \}$ and dimension $D\in\{2, 4, 8, \dots, 1024 \}$. In total, there are $100$ sets with varying dimension and standard deviation.
In order to evaluate the federated fuzzy $c$-means algorithm, we randomly (but uniformly) distribute the points among ten clients and run the federated clustering ten times for each dataset. That way, we have a benchmark for settings where the data is not unequally distributed and non-i.i.d.

Firstly, applying $avg_{1}$ or $avg_{2}$ leads to similar results in terms of ground truth knowledge gap. Given that the data is now uniformly distributed among clients, this is not a surprising result and consistent with observations in \cite{kumar_fed_kmeans_2020}. For summary statistics of the experiments, please refer to Table \ref{tab:g2_sets_gt}. The table summarizes the results of the knowledge gap metrics of the central clustering and $avg_{1}$, $avg_{2}$ clusterings. We see that the knowledge gap is almost the same.

\begin{figure}[t!]
    \centering
    \subfigure[]{\includegraphics[width=0.20\textwidth]{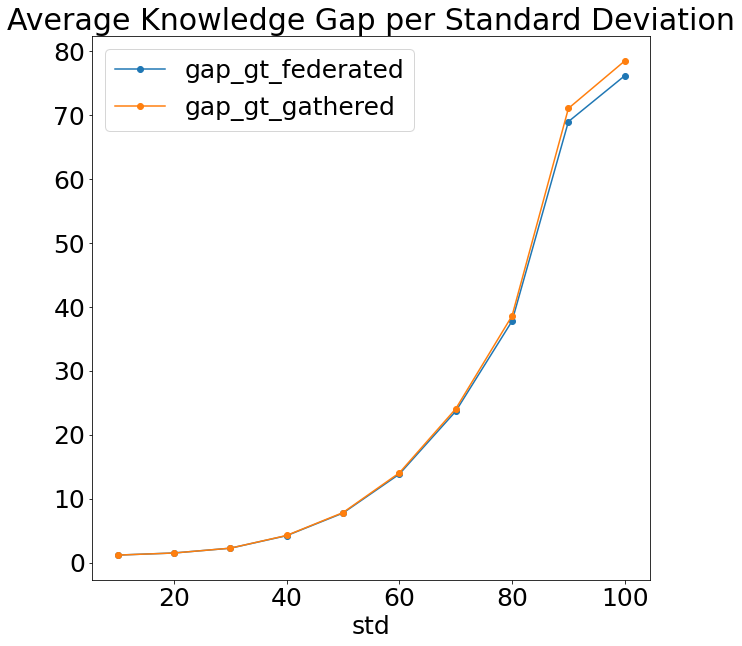}}
    \subfigure[]{\includegraphics[width=0.20\textwidth]{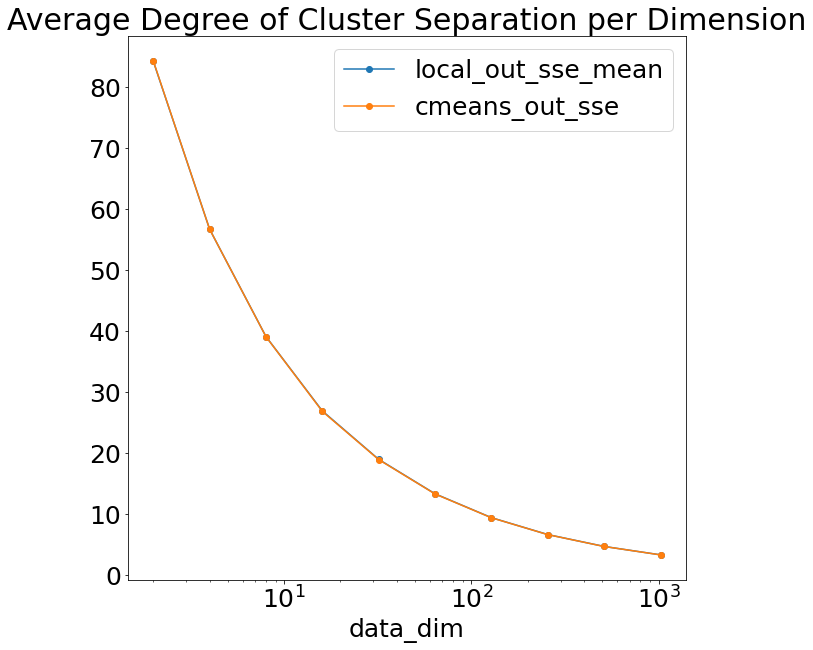}}
    \caption{Decreasing performance of federated and non-federated fuzzy $c$-means algorithms in the case 4 experiments(G2 sets) as measured by knowledge gap and outside cluster SSE. (a) The knowledge gap increases with higher standard deviation (= cluster overlap) on average. (b) The cluster separation decreases with higher dimensionality on average as measured by outside cluster SSE on the y-axis. The x-axis is scaled logarithmically.}
    \label{fig:g2_sets_weaknesses}
\end{figure}

Secondly, not only $avg_{1}$ and $avg_{2}$ produce similar results, but also the federated and non-federated version converge to similar cluster centers (Table \ref{tab:g2_sets_gt}) in all of our test sets. In particular, that means both versions have the same strengths and weaknesses. That is, they work well on low-dimensional ($\approx d<64$, relatively good cluster separation as indicated by outside SSE) data with small overlap ($\approx \sigma <60$, relatively small knowledge gap) and struggle with higher dimensional data and greater overlap. See Figure \ref{fig:g2_sets_weaknesses} for a visualization of the knowledge gap per standard deviation and SSE metric per dimension.

\begin{table}[t!]
\centering
\resizebox{0.30\textwidth}{!}{%
\begin{tabular}{l|l|ll}
 & Non-Federated & \multicolumn{2}{l}{Federated} \\ \hline
\begin{tabular}[c]{@{}l@{}}Knowledge Gap\\ Statistic\end{tabular} & \begin{tabular}[c]{@{}l@{}}Central\\ Clustering\end{tabular} & $avg_{1}$ & $avg_{2}$ \\ \hline
\begin{tabular}[c]{@{}l@{}}25\%-\\ quantile\end{tabular} & 2.42 & 2.43 & 2.41 \\ \hline
\begin{tabular}[c]{@{}l@{}}50\%-\\ quantile\end{tabular} & 9.01 & 8.87 & 8.94 \\ \hline
\begin{tabular}[c]{@{}l@{}}75\%-\\ quantile\end{tabular} & 30.98 & 30.33 & 30.38 \\ \hline
min & 0.98 & 0.98 & 0.97 \\ \hline
max & 99.80 & 99.24 & 95.55
\end{tabular}%
}
\caption{Ground truth gaps calculated in case 4 experiments (100 G2 test sets) for the non-federated (centrally clustered) algorithm as well as the federated algorithms ($avg_{1}$ and $avg_{2}$). All approaches result in similar clusters.}
\label{tab:g2_sets_gt}
\end{table}

\section{Conclusion}
With this work, we hope to motivate more efforts towards federated clustering. We introduce two application cases in which federated clustering can create value, describe a federated fuzzy $c$-means (FFCM) protocol with two alternative averaging functions, and evaluate their behaviours in challenging situations through numerical experiments. In particular, the results obtained by applying the $k$-means algorithm as averaging function are promising. Through this method, we are able to detect good global cluster centers even in challenging situations. However, it does come at a high computational cost.
Some challenges are not addressed in this work. For instance, improving the random initialization method in FFCM has not been investigated. Moreover, we assumed the number of global clusters to be known. The challenge of determining that number in a federated setting remains open. Also, the impact of $m$ was not subject to a principled evaluation. Further, we evaluate the approach on challenging, but purely artificial datasets. It is still to determine whether the approach is of significant practical value, e.g., through evaluation on ``real-world" datasets or through embedding the clustering algorithm into a supervised FL framework. Results from those experiments can then also be compared to other works. Finally, the evaluation of FFCM was purely empirical such that a more rigorous and theoretical study on convergence properties and behaviour still needs to be performed.

All in all, we see first promising results leading us to conclude that the approach is viable and open questions are worth to be studied in future works.

\section{Acknowledgement}
This work was supported by Vodafone GmbH.

\bibliography{aaai22.bib}

\end{document}